%% file: paper3477.tex
\documentclass[runningheads]{llncs}

\usepackage{graphicx}
\usepackage{tabularx}

\usepackage{amssymb}
\usepackage{amsbsy}

\begin{document}

\title{SG-Shuffle: Multi-aspect Shuffle Transformer for Scene Graph Generation}
 
\author{
Anh Duc Bui
\thanks{Corresponding author.}
\and
Soyeon Caren Han 
\and
Josiah Poon
}
%
%
\institute{The University of Sydney, Camperdown, Sydney, Australia \\
\email{abui2208@uni.sydney.edu.au, \{caren.han, josiah.poon\}@sydney.edu.au }\\
}
 
\maketitle               

\begin{abstract}
Scene Graph Generation (SGG) serves a comprehensive representation of the images for human understanding as well as visual understanding tasks. Due to the long tail bias problem of the object and predicate labels in the available annotated data, the scene graph generated from current methodologies can be biased toward common, non-informative relationship labels. Relationship can sometimes be non-mutually exclusive, which can be described from multiple perspectives like geometrical relationships or semantic relationships, making it even more challenging to predict the most suitable relationship label. In this work, we proposed the SG-Shuffle pipeline for scene graph generation with 3 components: 1) Parallel Transformer Encoder, which learns to predict object relationships in a more exclusive manner by grouping relationship labels into groups of similar purpose; 2) Shuffle Transformer, which learns to select the final relationship labels from the category-specific feature generated in the previous step; and 3) Weighted CE loss, used to alleviate the training bias caused by the imbalanced dataset.

\end{abstract}

\input{Introduction}

\input{RelatedWork}

\input{Methodology}

\input{Evaluation_Setup}

\input{Performance_Analysis}

\input{Conclusion}

\bibliographystyle{splncs04}

{\tiny
\bibliography{references}}

\end{document}

%% file: Introduction.tex
\section{Introduction}

Scene Graph Generation (SGG) is a fundamental visual understanding task that aims to encode image structure using the objects in the image as well as the relationships between these objects into a more compact representation with graphs \cite{johnson2015image}. Such representation allows for a more comprehensive understanding of the visual scene and serves as an intermediate data structure for downstream machine learning tasks between images and text, such as VQA \cite{luo2020rexup} or Text-Image Matching \cite{long2022gradual}. Significant progress has been made recently in SGG thanks to the advancement of object detection \cite{ren2015faster}. However, due to the challenges of variation in object-predicate type as well as the extremely long tail bias of objects and predicates, efforts for SGG must be made so that scene graphs can be more effective for other visual understanding tasks. The traditional pipeline of SGG can often be viewed as a design pattern that comprises 2 main parts, with the predicate prediction built on top of the object detector, which generates object feature representation through convolution neural network structure. Most methods focus on leveraging contextual object features in images via a variety of message propagation mechanisms such as LSTM \cite{li2017scene,tang2019learning,zellers2018neural,guo2021relation} and GNN \cite{yang2018graph,mi2020hierarchical,knyazev2020graph}. Such methods include the biassed prediction of predicate labels towards the head categories with much lower performance in tail categories. This is a major problem for the intended purpose of scene graphs; head categories frequently have generic meanings but tail categories provide important information that can be used in downstream tasks. Recent research has been conducted toward solving this long tail bias by a number of debiasing methods: data augmentation \cite{li2021bipartite,guo2021general}, model design \cite{yan2020pcpl,yu2020cogtree}, and bias disentangling \cite{tang2020unbiased,chiou2021recovering}. These methods focus on making use of predicates' frequency and the hierarchy structure of predicates' correlation with object labels to make their models focus on infrequent predicates. There has been a lack of research into non-correlated predicates, which are used for different purposes but might have a non-mutually exclusive distribution of objects and subjects. We argue that this leads to the problem where the model needs to give attention to the classification between predicate labels of different purposes and semantic features like ``above," which is used to describe positional relationships, and ``holding," which is used to describe an action. This leads to less focus on differentiating between predicate labels of similar purpose like ``above" and ``under", which is already challenging due to the long tail biased problem presented in the SGG task. The SGG model can learn the differences between predicates with similar or contrasting semantic correlations and reduce the bias of the tail class towards the head class of different semantic spaces.

To tackle the challenge, we propose the SG-Shuffle architecture that limits the learning of classification between predicate labels in different semantic spaces to improve classification between semantically correlated predicate labels. In order to separate non-correlated predicate labels, we group correlated predicates into four groups: Geometric, Possessive, Semantic, and Misc based on their purpose and super-type following the description in the Neural Motif paper \cite{zellers2018neural}. A stacked transformer encoder is adopted for feature refinement and contextual information encoding of the object feature to generate the category-specific predicate feature with fine-grained information that distinguishes predicate with correlated semantics. A shuffle transformer structure based on Transformer \cite{vaswani2017attention} and ShuffleNet \cite{ma2018shufflenet} is proposed to fuse such fine-grained category-specific features into a more universal feature that can classify between all predicates labels in the dataset. This structure both fuses the fine-grained features generated from the previous step and further propagates contextual information among the scene graphs. We then applied the simple loss weighting strategy at the end of the training process to further handle the long tail bias problem that also exists within the predicate of the same category.

Our contributions are as follows:
First, we addressed the SGG issue where uncorrelated labels are classified against each other, which we tackled by categorising correlated labels and learning category-specific predicate features.
Second, we also proposed a Shuffle Transformer layer, which is used to fuse features of different focuses to obtain the universal predicate feature for predicate classification as part of our architecture, SG-Shuffle.
Third, we evaluated the performance of the proposed SG-Shuffle to demonstrate its effectiveness in the SGG task.

%% file: RelatedWork.tex
\section{Related Works}
Scene graphs received an attention in vision and language joint learning research as they can serve as a structural representation of images and have the potential to benefit several downstream vision and language reasoning tasks, such as image generation, image retrieval, visual question answering, and image captioning. Earlier works in scene graph generation involve making better use of visual features. They leverage contextual information for object prediction and predicate prediction using message passing \cite{xu2017scene}, LSTMs \cite{li2017scene,tang2019learning,zellers2018neural,guo2021relation}, and GNN \cite{yang2018graph,mi2020hierarchical,knyazev2020graph}. Statistics correlation of object and predicate are also used in addition to give the models more information to enhance the results. \cite{zellers2018neural} has used GloVe for implicit statistics correlation, whereas \cite{chen2019knowledge} has explicitly used statistical correlation as edges in GNN. 
While performance was improved, challenges still remain due to the long-tailed data distribution which causes these models to perform poorly on infrequent classes. Recent work has looked at several debiasing methods for unbiased scene graph generations which can mainly categorised into three major types: re-sampling; loss re-weighting; and bias disentanglement from biased results. \cite{li2021bipartite} proposed to oversampling image instances while under-sampling common predicates for balanced predicate distribution. \cite{yan2020pcpl} and \cite{guo2021general} on other hand suggest to use label correlation to realign their training loss while other methods like \cite{suhail2021energy,zhang2019graphical,knyazev2020graph} propose their additional training loss objectives to reduce the bias problems. Other than re-sampling and loss re-weighting, bias disentanglement is also commonly used, removing bias from biased model result for unbiased scene graph. \cite{tang2020unbiased} propose to remove causal inference bias while missing label bias is estimated from label frequency and removed in \cite{chiou2021recovering}. 


One of the challenge in computer vision is the channel sparse connection problem in convolution neural networks for images, where each convolution only operates on a single group of input channels due to the use of group convolutions for reducing model complexity. ShuffleNet \cite{ma2018shufflenet} was proposed to address the problem by allowing for information exchange between channels of different groups through the use of channel shuffle operations between group convolution layers. Inspired by this, channel shuffling was used in multiple works in a variety of different deep learning research works \cite{wang2020learning,geng2021dynamic} to allow for information flow and strengthen the correlation between components of their model. Based on the success of ShuffleNet, we propose Shuffle Transformer containing the channel shuffling operation for combining multiple category specific predicate features.


%% file: Methodology.tex
\section{Methodology}
The typical methods of SGG comprise a two-stage process: 1) detecting objects within the images and 2) predicting the relationships between these objects. In the first stage, a standard object detector like Faster RCNN \cite{ren2015faster} obtains a set of bounding boxes for the set of objects detected in the image. RoIAlign \cite{ren2015faster} generates the visual feature of these bounding boxes and determines the initial detection of the object label for each of the detected objects. The object bounding boxes, which represent the position of the object in the image; object visual features, which represent the shape, form, and pattern learned by the object detector about the object; and object labels, which represent natural language understanding of the object semantic, are predicted using the input image and used as input for the next step. If ground truth information is used, as is the case of PredCls or SGCls settings, the ground truth information is inserted at the step where the information is intended to be used. In the second stage, the information generated from the object detector is used to predict the predicate between the predicted objects. As Faster RCNN is usually used for object detection, SGG models generally focus on the second stage of the process, which is also aligned with the main focus of our proposed SG-Shuffle. Our proposed model for predicate prediction, in particular, consists of three steps: 1) Four individual transformer sub-models are used to learn the category-specific representation of the objects and predicates; 2) Shuffle Transformer layers are then used to merge and allow information flow between the previous step's output; and 3) Finally, weighted cross-entropy (CE) loss is calculated and used for model optimization as a way to reduce the long tail biased problem.

\subsection{Categories}
To focus the attention of the model on distinguishing between predicate labels of similar semantic space, we categorize the predicate labels into 4 groups based on their super-type following the description of the Neural Motif\cite{zellers2018neural}: Geometric, Possessive, Semantic, and Misc as shown in Table 1. We limit the need of the model to classify between predicates with different semantic purposes which can often require attention to different aspects of the input, for example: mainly object position for Geometric predicates, object label, and visual feature for Possessive predicates, or a balanced combination of the three for Semantic and Misc predicates. And this, in turn, allows the model to make use of the input aspects selectively to classify between semantically correlated predicates of the same category, which can be challenging for general scene graph models since they are often represented close to each other in the feature space due to semantic similarity, especially with the long tail biased problem of the SGG dataset.

\begin{table}[h]
\caption{Predicate Categories and predicate labels in each category}

\centering
\begin{tabularx}{\textwidth}{|c|X|} 

\hline
\textbf{Category}   & \textbf{Predicate label}                                                                                                                                                                                                                                                                                      \\ 
\hline
Geometric  &`above’, `across’, `against’, `along’, `at’, `behind’, `between’, `in front of’, `near’, `on’, `on back of’, `over’, `under’, `in’ , `and'                                                                                                                                                             \\ 
\hline
Possessive &`belonging to’, `has’, `part of’, `wearing’, `attached to’, `of’, `wears’, `with`                                                                                                                                                                                                                     \\ 
\hline
Semantic   & `to’, `carrying’, `covered in’, `covering’, `eating’, `flying in’, `growing on’, `hanging from’, `holding’, `laying on’, `looking at’, `lying on’, `mounted on’, `painted on’, `parked on’, `playing’, `riding’, `says’, `sitting on’, `standing on’, `using’, `walking in’, `walking on’, `watching'  \\ 
\hline
Misc       &`for’, `from’, `made of'                                                                                                                                                                                                                                                                              \\
\hline
\end{tabularx}

\end{table}

\subsection{Parallel Transformer Encoder}

For each image, bounding boxes, corresponding object features, and object labels are generated using Faster RCNN \cite{ren2015faster} as input to our relationship prediction module. In order to incorporate object label information as input, object labels are encoded using GloVe encoding \cite{pennington2014glove}. As these inputs are still non-contextual and are not specifically trained for relationship prediction, as shown in Figure 1, we made use of the transformer encoder architecture to transverse contextual information as well as refine the feature vectors for relationship prediction. And since we needed 4 sub-models to learn the specific details about the relationship categories, the 4 transformer encoders are trained with their own goal of classifying relationships within each of the category groups. For simplicity, we concatenate the three object information streams, including object bounding box, object feature, and object label encoding, as input to our context encoder.
\begin{equation}
Input= W_{o} [pos(box_{i}), visual_{i}, GloVe(label_{i})]  
\end{equation}
Our context encoder adopts the out-of-the-box architecture of the transformer encoder as it has been shown to be relatively effective compare to RNN, or CNN in both natural language and computer vision. It is composed of layers of self-attention, feed-forward, and layer normalization stacked.
\begin{equation}
o'_{c} =  LayerNorm(SelfAttention(o_{c}) + o_{c})
\end{equation}
\begin{equation}
o’{}’_{c} = LayerNorm(FeedForward(o'_{c}) + o'_{c})
\end{equation}
where o\textsubscript{c} is the input object feature o'\textsubscript{c} is the output of the multi-head attention and 
o'{}'\textsubscript{c} is the output of the transformer encoder.
The output of the contextual encoder is combined with the bounding boxes and visual features of the unions and intersections through concatenation to predict the category-specific relationships, which calculates a category-specific loss using CE loss.
\begin{equation}
r_{c} = Softmax( Concat(subject_{c},object_{c},v_{intersect},v_{union}))
\end{equation}
where r\textsubscript{c} is the category specific predicate features, v is the visual feature while subject\textsubscript{c} and object\textsubscript{c} are the output object features of the objects pair.

We applied CE loss to jointly optimise the sub-models in their respective categories. The parameters of the sub-models are trained in parallel and optimized as part of the training process. For each of the sub-models, object pairs with relationships from other categories are not considered in the computation of this loss.

\begin{figure*}[ht]
  \centering
  \includegraphics[width=\linewidth]{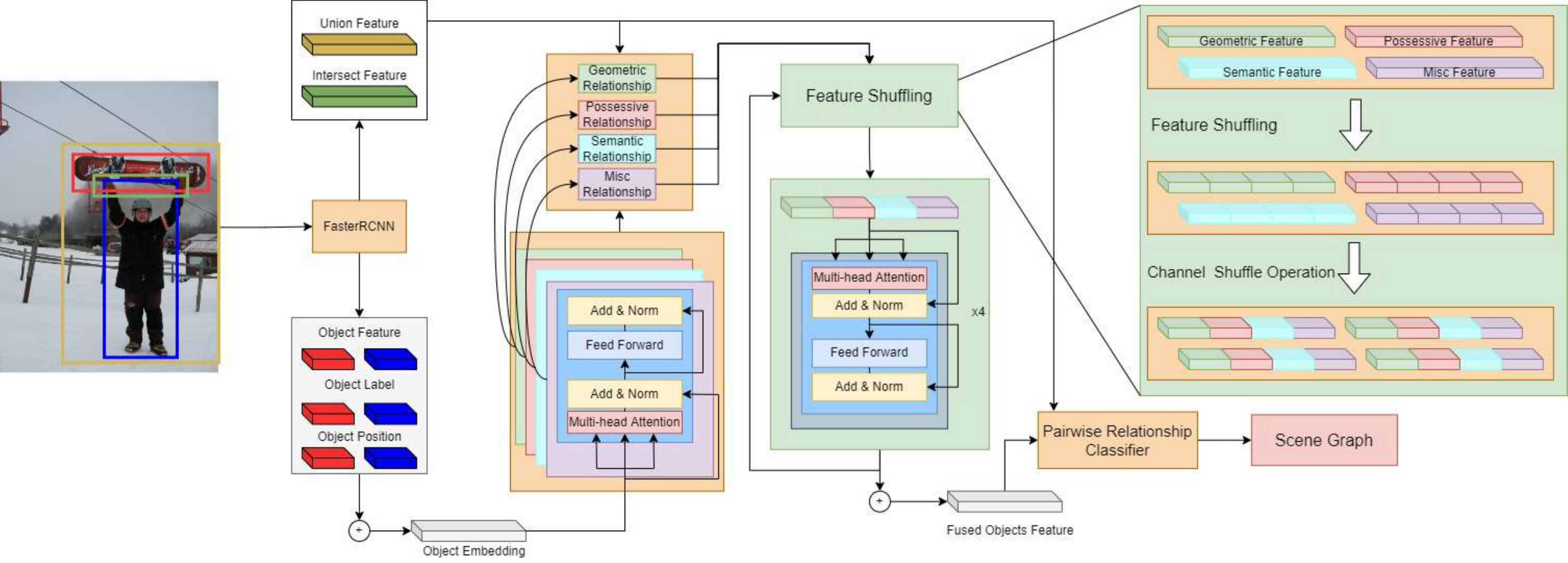}
  \caption{Overall Architecture of SG-Shuffle}
\end{figure*}

\subsection{Shuffle Transformer}

In the second step of our relationship prediction pipeline, after the sub-model learns the category-specific contextual information for each of the objects in the image, these category-specific contextual object features are merged together in order to classify the actual relationship of the objects pair in the original relationship labels set. Therefore, in this stage, the outputs of the transformer encoder sub-models are used. Furthermore, while the 4 categories used in the previous step are from different semantic domains, they are still correlated as the candidate relationship of the same pair of objects, and hence, information flow between these sub-models is needed. In order for the aforementioned reasons to be incorporated into the model, we need to ensure that all the outputs of the sub-models are relevant in the prediction stage and the correlation between these category-specific object features is taken into account to further improve the prediction result. We proposed making use of the shuffle architecture, which was proposed in ShuffleNet \cite{ma2018shufflenet} to handle information flow between channels of CNN for computer vision tasks for this purpose. As shown in Figure 1, this architecture specifically makes use of the channel shuffle operation to allow such information flow. The original ShuffleNet is used with CNN for images and is not directly applicable to our situation, so we replace their convolution layer with a transformer encoder layer with the same architecture as in the previous step, with shuffle layers in between similar to the shuffle net. For SG-Shuffle, since we have 4 category-specific object features of 4 sub-models from the previous stage, for simplicity, the same number of shuffle sub-models are used in our shuffling stage. By using four shuffle sub-models, a quarter of the output features from the previous layer are concatenated to be the input of the next layer. 
\begin{equation}
o'_{s} = SelfAttention([partition_{s}(o_{(k=1\longrightarrow 4)})])
\end{equation}
\begin{equation}
o_{final} = W_{s}[o^{final-1}_{(s=1\longrightarrow 4)}]
\end{equation}
\begin{equation}
r_{final} = Softmax(Concat(subject_{final},object_{final},v_{intersect},v_{union}))
\end{equation}
After a few shuffle layers, the output of the shuffle sub models are concatenated and used to predict the predicate of the object pairs using the softmax function.

\subsection{Weighted CE Loss}
While categorizing the predicate labels into 4 different groups helps alleviate the bias problem to a certain degree, there is still bias between predicates of the same predicate group. To further remedy the long tail bias problem in the SGG, at the end of the training process, we applied a simple re-weighted CE loss to balance the learning process of each predicate label. Traditionally, for classification tasks like predicate prediction in SGG, a network is trained to minimise the CE loss. The predicted probability is obtained by applying the Softmax function to the output of the final layer. This loss penalises errors of predicting each label equally and therefore makes the model skew toward common labels due to the number of instances they have in the dataset. The weighted CE loss is a simple modified version of this CE loss with larger weights for the infrequent labels and lower weights for the frequent labels and penalises error classification accordingly.
\begin{equation}
l(x,y) = L  \{l_{1},...,l_{N} \}^{T} , l_{n} = -\sum_{c=1}^{C}w_{c}log\frac{exp(x_{n,c})}{\sum_{i=1}^{C}exp(x_{n,i})}y_{n,c}
\end{equation}

%% file: Evaluation_Setup.tex
\section{Evaluation Setup}
\textbf{Dataset Details} We used the VG150 dataset \cite{zellers2018neural}, a subset of the large-scale Visual Genome vision and language dataset. It is the pre-processed split which is specifically used for SGG tasks, with the most frequent 150 object categories and 50 predicates categories. For object and predicate-super categories, we followed \cite{zellers2018neural} criteria to split the predicate classes into 4 super-classes based on their semantic nature. Following the same testing strategy, as \cite{tang2020unbiased}, we also use the original split with 70\% training set and 30\% test set, as well as taking 5000 samples from the training set as a validation set for parameter adjustment.

\noindent
\textbf{Evaluation Metrics} We use the mean Recall@ K metric. This metric has recently been used in place of regular recall due to the long tail bias problem in the image dataset, which leads to the performance bias in this metric \cite{tang2020unbiased}. The evaluation is done by predicting the relationship triplets in 3 settings: Predicate Classification (PredCls): using the image with ground truth object label and bounding box, Scene Graph Classification (SGCls): only ground truth bounding box and Scene Graph Detection (SGDet): using only the ground truth image.

\noindent
\textbf{Implementation Details} We use the Faster RCNN as the object detector to focus on the performance of the predicate prediction and stay consistent with previous work. It is pre-trained on ImageNet and fine-tuned on VG150 by \cite{tang2020unbiased} with ResNeXt-101-FPN being the backbone for region proposals. For consistency with previous works, the parameters of the object detector were kept frozen during the training and evaluation period. The stacked encoder used as the category sub-models contains 6 layers of transformer encoder with 4 attention heads each. For the weighted CE loss, we applied the inversed square root of predicate frequency as mentioned in \cite{chiou2021recovering} as weight for the loss function. We optimised the proposed model using the Adam optimizer with an initial learning rate of 0.001 and the warm-up and decay strategy suggested by \cite{tang2020unbiased}. The experiment was conducted on the NVIDIA T4 GPU.

%% file: Performance_Analysis.tex
\section{Performance Analysis}
\subsection{Quantitative Evaluation}
We compare SG-Shuffle with other SGG methods to demonstrate the ability of the proposed SG-Shuffle architecture to improve upon the feature refinement of objects and relationships in SGG while also displaying that it can be used with debiasing methods for unbiased SGG.

\begin{table}[]

\def\arraystretch{1.25}%
\centering

\caption{Performance Evaluation on VG150.}

\resizebox{\textwidth}{!}{%
\begin{tabular}{|ll|cc|cc|cc|}
\hline
\multicolumn{2}{|l|}{{}{}{}}                   & \multicolumn{2}{c|}{\textbf{PredCLS}}                             & \multicolumn{2}{c|}{\textbf{SGCLS}}                               & \multicolumn{2}{c|}{\textbf{SGDET}}                               \\ \cline{3-8} 
\multicolumn{2}{|l|}{}                                            & \multicolumn{1}{c|}{\textbf{mR@50}}            & \textbf{mR@100 }          & \multicolumn{1}{c|}{\textbf{mR@50}}            & \textbf{mR@100 }          & \multicolumn{1}{c|}{\textbf{mR@50}}            &\textbf{ mR@100}           \\ \hline
\multicolumn{1}{|l|}{\textbf{Without}}   & IMP (2017)                      & \multicolumn{1}{c|}{9.80\%}           & 10.50\%          & \multicolumn{1}{c|}{5.80\%}           & 6.00\%           & \multicolumn{1}{c|}{3.80\%}           & 4.80\%           \\ \cline{2-8} 
\multicolumn{1}{|l|}{\textbf{Debiasing}} & Motif (2018)                    & \multicolumn{1}{c|}{13.30\%}          & 14.40\%          & \multicolumn{1}{c|}{7.10\%}           & 7.60\%           & \multicolumn{1}{c|}{5.30\%}           & 6.10\%           \\ \cline{2-8} 
\multicolumn{1}{|l|}{}          & KERN (2019)                     & \multicolumn{1}{c|}{17.70\%}          & 19.20\%          & \multicolumn{1}{c|}{9.40\%}           & 10.00\%          & \multicolumn{1}{c|}{6.40\%}           & 7.30\%           \\ \cline{2-8} 
\multicolumn{1}{|l|}{}          & VCTree (2019)                   & \multicolumn{1}{c|}{{ 17.90\%}}    & { 19.40\%}    & \multicolumn{1}{c|}{{ 10.10\%}}    & { 10.80\%}    & \multicolumn{1}{c|}{{ 6.90\%}}     & { 8.00\%}     \\ \cline{2-8} 
\multicolumn{1}{|l|}{}          & Our model                       & \multicolumn{1}{c|}{\textbf{24.39\%}} & \textbf{25.94\%} & \multicolumn{1}{c|}{\textbf{13.00\%}} & \textbf{13.90\%} & \multicolumn{1}{c|}{\textbf{10.94\%}} & \textbf{12.01\%} \\ \hline
\multicolumn{1}{|l|}{\textbf{With}}      & Motif + TDE (2020)              & \multicolumn{1}{c|}{25.50\%}          & 29.10\%          & \multicolumn{1}{c|}{13.10\%}          & 14.90\%          & \multicolumn{1}{c|}{8.20\%}           & 9.80\%           \\ \cline{2-8} 
\multicolumn{1}{|l|}{\textbf{Debiasing}} & PCPL (2020)                     & \multicolumn{1}{c|}{{ 35.20\%}}    & { 37.80\%}    & \multicolumn{1}{c|}{\textbf{18.60\%}} & \textbf{19.60\%} & \multicolumn{1}{c|}{9.50\%}           & 11.70\%          \\ \cline{2-8} 
\multicolumn{1}{|l|}{}          & Motif + DLFE (2021)             & \multicolumn{1}{c|}{26.90\%}          & 28.80\%          & \multicolumn{1}{c|}{15.20\%}          & 15.90\%          & \multicolumn{1}{c|}{11.70\%} & 13.80\% \\ \cline{2-8} 
\multicolumn{1}{|l|}{}          & BGNN (2021)                     & \multicolumn{1}{c|}{30.40\%}          & 32.90\%          & \multicolumn{1}{c|}{14.30\%}          & 16.50\%          & \multicolumn{1}{c|}{10.70\%}          & 12.60\%          \\ \cline{2-8} 
\multicolumn{1}{|l|}{}          & Our Model/w Weighted Loss & \multicolumn{1}{c|}{\textbf{35.57\%}} & \textbf{38.67\%} & \multicolumn{1}{c|}{17.96\%} & 19.24\%    & \multicolumn{1}{c|}{\textbf{13.52\%}} & \textbf{14.91\%}   \\ \hline
\end{tabular}
}

\label{tab:my-table}
\end{table}

Firstly, we compare SG-Shuffle without weighted CE loss to other biased SGG baselines, including IMP \cite{xu2017scene}, Motif \cite{zellers2018neural}, KERN \cite{chen2019knowledge}, VCTree \cite{tang2019learning}. These models aims to generate better objects feature representations by traversing context information between objects in the images. We compare SG-Shuffle without weighted CE with these baselines to demonstrate the effectiveness of SG-Shuffle in generating informative feature representation. As shown in the first part of Table 2, the mR@100 of our model is 6.5\% higher in the PredCLS setting, 3.1\% higher in the SGClS setting, and 4.0\% higher in the SGDet setting comparing to the VCTree, which is the best performing model among models without debiasing methods. These models worked on the large label set at once, a challenging task since the distance between the predicate label are not uniform in the feature space. SG-Shuffle was able to gain better performance by learning in-depth features that differentiate predicate with close semantic nature.

Secondly, we compare SG-Shuffle with weighted CE loss with the more recent unbiased SGG models such as TDE \cite{tang2020unbiased}, PCPL \cite{yan2020pcpl}, DLFE \cite{chiou2021recovering}, and BGNN \cite{li2021bipartite}, which use debiasing strategies to solve the long-tailed bias problem in the SGG task. We observed that with simple weighted CE loss, SG-Shuffle outperforms baseline models in the PredCls setting and SGDet setting, in which our PredCls score is 0.87\% higher than PCPL and 5.77\% higher than DLFE in mR@100 score. It only comes slightly lower than only PCPL in SGCls by a minor 0.3\%. Among the baseline models, strategy used by PCPL also involve learning a better representation of predicates by modeling the relationship between predicate labels. Compare to our model and PCPL, other models in the baseline are designed to reduce the training bias by removing biased probability or re-sampling, to outperform models without debiasing. But without in-depth learning of predicate representation, their performance is generally lower than the models with this feature like PCPL and our model.

\subsection{Hyper Parameter Tuning}
While increasing the number of layers is often advantageous in the early layers of deep learning models, at a higher number of layers, it could also lead to diminishing gradient and optimization issues. We conducted hyper-parameter testing with it being tested with a varying number of shuffling layers. 

\begin{table}[ht]
\caption{Performance with shuffle layers in PredCLS, SGCLS, and SGDET.}

\centering
\def\arraystretch{1.25}%
\resizebox{.8\textwidth}{!}{%
\begin{tabular}{|c|cc|cc|cc|}
\hline
\textbf{\# of Shuffle} & \multicolumn{2}{c|}{\textbf{PredCLS}}                             & \multicolumn{2}{c|}{\textbf{SGCLS}}                               & \multicolumn{2}{c|}{\textbf{SGDET}}                               \\ \cline{2-7} 
\textbf{Layers }   & \multicolumn{1}{c|}{\textbf{mR@50}}            & \textbf{mR@100   }        & \multicolumn{1}{c|}{\textbf{mR@50}}            &\textbf{ mR@100 }          & \multicolumn{1}{c|}{\textbf{mR@50}}            & \textbf{mR@100}           \\ \hline
4         & \multicolumn{1}{c|}{30.00\%}          & 32.43\%          & \multicolumn{1}{c|}{14.72\%}          & 15.87\%          & \multicolumn{1}{c|}{10.59\%}          & 11.90\%          \\ \hline
5         & \multicolumn{1}{c|}{35.09\%}          & 37.68\%          & \multicolumn{1}{c|}{\textbf{17.96\%}} & \textbf{19.24\%} & \multicolumn{1}{c|}{\textbf{13.52\%}} & \textbf{14.91\%} \\ \hline
6         & \multicolumn{1}{c|}{\textbf{35.57\%}} & \textbf{38.67\%} & \multicolumn{1}{c|}{16.64\%}          & 17.81\%          & \multicolumn{1}{c|}{11.46\%}          & 12.78\%          \\ \hline
7         & \multicolumn{1}{c|}{32.13\%}          & 35.04\%          & \multicolumn{1}{c|}{14.89\%}          & 16.03\%          & \multicolumn{1}{c|}{9.92\%}           & 11.23\%          \\ \hline
\end{tabular}%
}

\label{tab:my-table}

\end{table}
In Table 3, we tested the model with 4, 5, 6, and 7 layers of shuffled transformer in all three SGG settings: PredCls, SGCls, and SGDet, and compare the performance using the mR@100 and mR@50 metrics. As shown in the table, the performance of the model increases significantly when the number of shuffle layers goes from 4 to 5 and goes down from 6 to 7. At this depth, challenge in optimization outweighs the performance gain of further layer depth increase. The model performs best with the PredCls setting when using 6 layers of shuffling, while 5-layered models perform best in the SGCls and SGDet settings. 

\subsection{Alternate Shuffle Layer}
In order to learn how the level of connection between the sub-models in the shuffling layer of the model affects the final performance, we also tested a pair-to-pair shuffling layer which is shown in Figure 2. In this setting, the output features from the previous layer are each partitioned into halves and each half is combined with a half from a different sub models and used as input for the current layer. After a few layers of shuffling, the category specific context information is shared in all 4 sub-models pathways. As shown in Table 4, while the pair to pair shuffling procedure does increase performance when compared to the model without any shuffling by 2\% in PredCls setting and 5\% in SGDet setting, the performance increase was lower than the full channel shuffling. We attribute the higher performance of the full shuffling layer over the pair-to-pair shuffling layer to the direct connection with all 4 sub-model from previous layer that allows it to learn important aspects from the previous layers at a faster rate and give better feature representation for relationship prediction.

\begin{table}[t]
\caption{Performance of SG-Shuffle with full shuffle layers and pairwise shuffle layers}
\def\arraystretch{1.5}%
\centering
\resizebox{.9\textwidth}{!}{
\begin{tabular}{|c|cc|cc|cc|}
\hline
                     & \multicolumn{2}{c|}{\textbf{PredCLS}}           & \multicolumn{2}{c|}{\textbf{SGCLS}}             & \multicolumn{2}{c|}{\textbf{SGDET}}             \\ \cline{2-7} 
                     & \multicolumn{1}{c|}{\textbf{mR@50}}   & \textbf{mR@100 } & \multicolumn{1}{c|}{\textbf{mR@50}}   & \textbf{mR@100}  & \multicolumn{1}{c|}{\textbf{mR@50}}   & \textbf{mR@100  }\\ \hline
No Shuffle        & \multicolumn{1}{c|}{27.53\%}          & 29.85\%          & \multicolumn{1}{c|}{14.94\%}          & 16.32\%          & \multicolumn{1}{c|}{6.13\%}           & 7.54\%           \\ \hline

Pair-to-pair Shuffle & \multicolumn{1}{c|}{29.23\%} & 31.62\% & \multicolumn{1}{c|}{13.93\%} & 15.14\% & \multicolumn{1}{c|}{10.89\%} & 12.18\% \\ \hline
Full Shuffle         & \multicolumn{1}{c|}{\textbf{35.57\%}} & \textbf{38.67\%} & \multicolumn{1}{c|}{\textbf{16.64\%}} & \textbf{17.81\%} & \multicolumn{1}{c|}{\textbf{11.46\%}} & \textbf{12.78\%} \\ \hline
\end{tabular}%
}
\label{tab:my-table}
\end{table}

\begin{figure}[h]
  \centering
  \includegraphics[width=.6\linewidth]{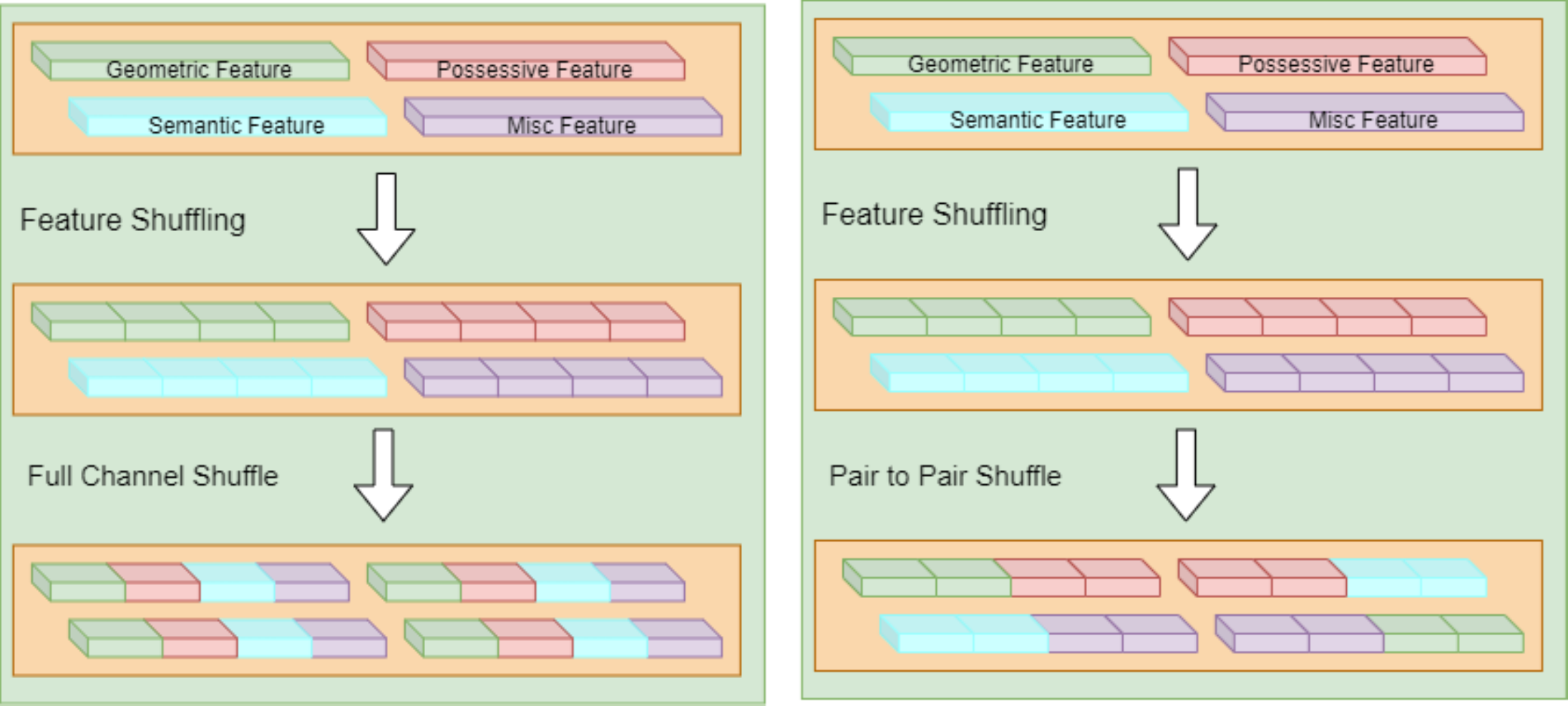}
  \caption{Full Channel Shuffling and Pair to Pair shuffling layer}
\end{figure}

\subsection{Ablation Study}
\subsubsection{Ablation Testing}
We performed an ablation study on our model by removing the shuffling layer or weighted CE, and comparing them with the full model to show the effectiveness of each component. Compare with the model without weighted CE loss, the full model has nearly a 13\% increase in performance in the PredCls setting when using the mR@100 metric. Long-tail bias plays a major part in SGG, and with no debiasing methodology, models are generally highly affected by the training bias introduced by the highly imbalanced dataset. 
\begin{table}[ht]
\centering
\caption{Ablation Result of SG-Shuffle with mR@100 and mR@50}
\def\arraystretch{1.5}%
\resizebox{.8\textwidth}{!}{%
\begin{tabular}{|c|c|cc|cc|cc|}
\hline
{}{}{\textbf{Shuffle}} & \textbf{Weighted} & \multicolumn{2}{c|}{\textbf{PredCLS}}                             & \multicolumn{2}{c|}{\textbf{SGCLS}}                               & \multicolumn{2}{c|}{\textbf{SGDET}}                               \\ \cline{3-8} 
                         & \textbf{CE} \textbf{Loss}  & \multicolumn{1}{l|}{\textbf{mR@50}}            & \textbf{mR@100}           & \multicolumn{1}{l|}{\textbf{mR@50}}            &\textbf{ mR@100}           & \multicolumn{1}{l|}{\textbf{mR@50}}            &\textbf{ mR@100    }       \\ \hline
\checkmark                        & -        & \multicolumn{1}{r|}{24.39\%}          & 25.94\%          & \multicolumn{1}{l|}{13.00\%}          & 13.90\%          & \multicolumn{1}{l|}{10.94\%}          & 12.01\%          \\ \hline
-                        & \checkmark        & \multicolumn{1}{r|}{27.53\%}          & 29.85\%          & \multicolumn{1}{l|}{14.94\%}          & 16.32\%          & \multicolumn{1}{l|}{6.13\%}           & 7.54\%           \\ \hline
\checkmark                        & \checkmark        & \multicolumn{1}{r|}{\textbf{35.57\%}} & \textbf{38.67\%} & \multicolumn{1}{l|}{\textbf{16.64\%}} & \textbf{17.81\%} & \multicolumn{1}{l|}{\textbf{11.46\%}} & \textbf{12.78\%} \\ \hline
\end{tabular}%
}

\label{tab:my-table}
\end{table}

When debiasing is included, the weighted CE loss model also has much lower performance than the full model by around 8\% mR@100 in the PredCls setting. This is due to the availability of shuffle layers, which allow information to flow more freely between the sub-models in the full model. Comparing the shuffled only model and the weighted CE loss only model, the weighted CE loss has the advantage of debiasing and has higher performance in PredCls and SGCls but loses out in the SGDet setting in which object position was omitted. This omission leads the model to rely on its own predicted object position for SGG, which is less reliable than ground-truth information and harder to refine without the use of shuffle layer. As shown in Table 5, both of the components of the SG-Shuffle are necessary to achieve higher performance in unbiased SGG.

\subsubsection{Categories Breakdown}

In Table 6, we look at the effect of the model component with respect to each category of predicate label by comparing the PredCls mR@100 of the models in each of the 4 categories. 
For the geometric category, the full model has the highest performance, while the weighted CE loss only model has only slightly higher performance than the shuffle only model. Since there are both common and uncommon relationships in this category, so the debiasing advantage of the weighted CE loss in uncommon classes is matched by the shuffle-only model, which performs better in common classes. In the possessive category, which is dominated by the common predicate classes the shuffle-only model has a slightly higher performance than the other two models.

\begin{table}[ht]
\caption{Effect on different categories using PredCLS setting with mR@100}

\def\arraystretch{1.25}%
\centering
\resizebox{.8\textwidth}{!}{%
\begin{tabular}{|c|c|c|c|c|c|c|}
\hline
{}{}{\textbf{Shuffle}} & \textbf{Weighted CE Loss} & {}{}{\textbf{Geometric}} & {}{}{\textbf{Possessive}} & {}{}{\textbf{Semantic}} & {}{}{\textbf{Misc}} & {}{}{\textbf{Overall}} \\ \hline
\checkmark                        & -        & 21.27\%                    & \textbf{31.33\%}            & 29.71\%                   & 4.73\%                & 25.94\%                  \\ \hline
-                        & \checkmark        & 22.06\%                    & 30.10\%                     & 37.16\%                   & 9.66\%                & 29.85\%                  \\ \hline
\checkmark                        & \checkmark        & \textbf{28.11\%}           & 31.11\%                     & \textbf{48.97\%}          & \textbf{29.29\%}      & \textbf{38.67\%}         \\ \hline
\end{tabular}%
}

\label{tab:my-table}
\end{table}

The main difference in the overall mR@100 lies in the semantic category with the largest number of predicates, which are both informative and comparably uncommon in the dataset. There is a large gap between the three models' performances in this category. The shuffle-only model suffered from the long-tailed bias, which affects the uncommon predicate in this category and performs the lowest of the three models. Between the full model and the weighted CE loss only model, the full model has the advantage of the shuffle transformer with more informative feature representation and much higher performance in the semantic category than the other two ablations tested models. Similarly, the mR@100 of the full SG-Shuffle model also has higher performance in the Misc category, which has uncommon predicate labels, than the ablation-tested models.

\subsection{Qualitative Analysis and Case study}
We visualize several scene graphs generated in the PredCls setting using the Ablation Tested Models in Figure 3. We selected 4 samples for the case study based on the objects presented in the images: A person's portrait; B) a person in a large background; C) a building; and D) animal with plants in the background.

\begin{figure}[ht]
  \centering
  \includegraphics[width=0.8\linewidth]{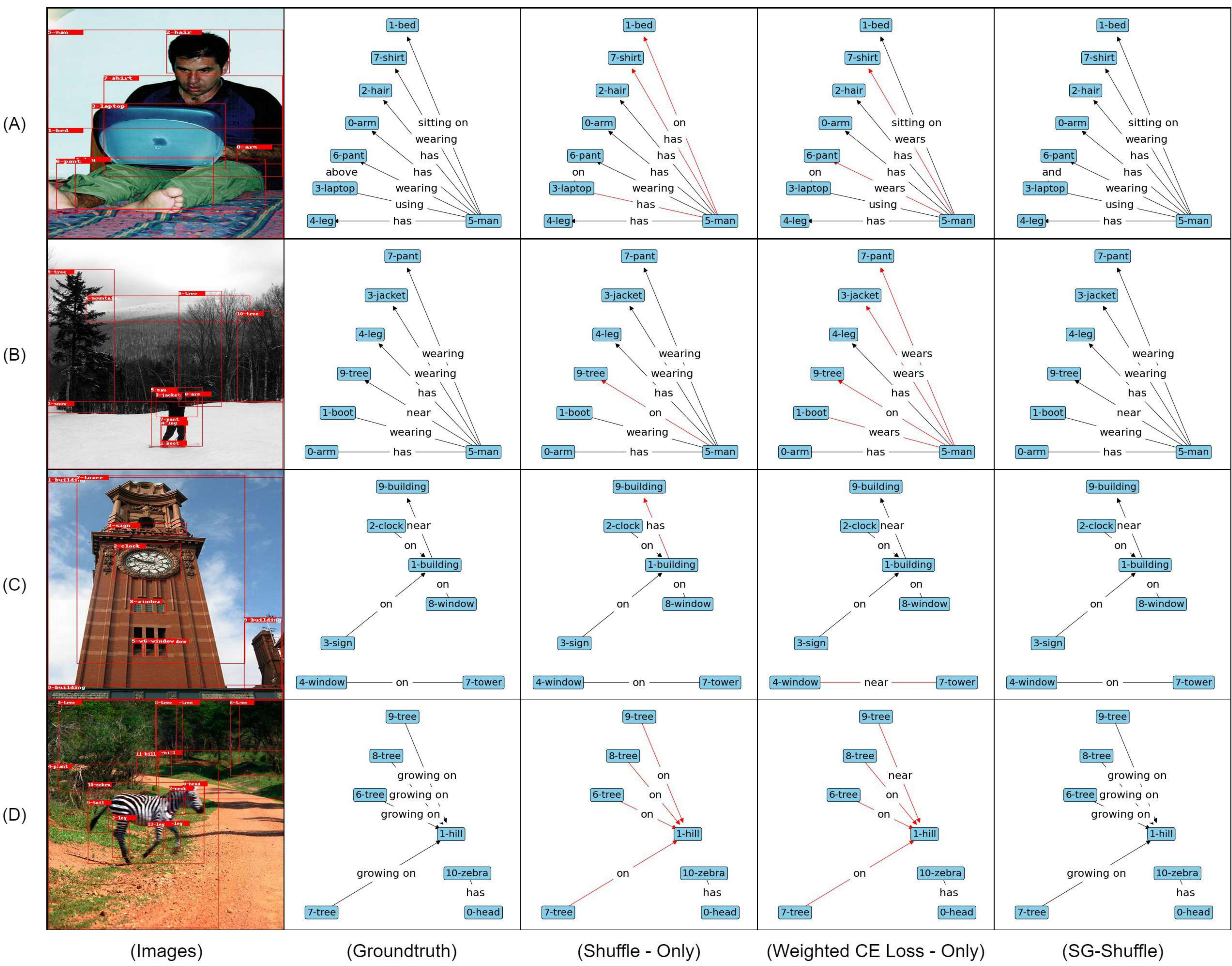}
  \caption{Sample Scene Graph generated from ablation tested models. Correct relationship is marked with black arrows and incorrect relationship is marked with red arrows}
\end{figure}

Observing samples A and B, between three models, the shuffle only model prefers more common predicates like ``on" or ``has", but cannot predict less frequent predicates like geometric relationship ``near", or semantic relationship ``using". The long-tailed bias present in the dataset to heavily affect the prediction of the model. The weighted CE loss only model, on the other hand, favors the infrequent relationship ``wear" over the more common but same meaning ``wearing". The SG-Shuffle model perform better than the other two models in both common possessive relationships like ``has" as well as infrequent semantic relationships like ``sitting on" or ``using". Similarly, in sample C, the shuffle only model misclassified infrequent relationship ``near", the weighted CE loss only does the opposite, misclassified common relationship ``on", while the SG-Shuffle correctly classifies both. However, weighted CE loss could not predict the every infrequent semantic relationship as shown in sample D, where ``growing on" was misclassified as "on". Comparatively, the full model was able to associate "growing on" with object ``tree" thanks to the improved feature representation from jointly learning from category-specific object features.

%% file: Conclusion.tex
\section{Conclusion}
In this paper, we propose the SG-Shuffle model for unbiased SGG by addressing non correlation problem of relationship labels in the existing SGG dataset. We proposed to categorise the set of predicate labels to four category ``Geometric", ``Possessive", ``Semantic", and ``Misc" in a divide and conquer approach which is learned as part of the SG-Shuffle SGG pipeline by Transformer Encoder in parallel to provide category-specific context for further SGG. We also propose a shuffle transformer layer, which apply channel wise shuffling operation in combination with the Transformer Encoder architecture to allow information flow between sub-models and merge together the learned category-specific feature representation. We demonstrated the effectiveness of SG-Shuffle in the VG150 dataset in comparison with other state-of-the-art SGG models.